\documentclass[conference]{IEEEtran}

\IEEEoverridecommandlockouts
\usepackage{float}
\usepackage{hyperref}

\usepackage[percent]{overpic}

\usepackage{gensymb}
\usepackage{amsfonts}
\usepackage{amsmath}
\usepackage{amssymb}
\usepackage{cite}
\usepackage{comment}
\usepackage[T1]{fontenc}
\usepackage{graphicx}
\usepackage[super]{nth}
\usepackage{textcomp}
\usepackage[font=smaller]{caption}
\usepackage{subcaption}
\usepackage{algorithm}
\usepackage{algcompatible}

\usepackage{color, colortbl}
\definecolor{LightGray}{gray}{0.85}
\definecolor{LightCyan}{rgb}{0.88,1,1}

\usepackage{todonotes}

\usepackage{balance}
\usepackage{xcolor}
\usepackage[export]{adjustbox}
\usepackage{titlesec}
\titlespacing{\section}{0pt}{*0.25}{*0.25}
\titlespacing{\subsection}{0pt}{*0.25}{*0.25}
\usepackage{geometry}
\begin{document}

\newgeometry{
 letterpaper,
 top=72pt,
 left=54pt,
 right=54pt,
 bottom=54pt,
 }

%\title{\LARGE \bf Robotic manipulation via dragging and pivoting using multi-modal contact state estimation}
%\title{\LARGE \bf Pivoting objects via in-hand manipulation using multi-modal sensing}
%\title{\LARGE \bf Pivoting Objects via In-Hand Manipulation using Vision, Force and Touch}
\title{\LARGE \bf Rotating Objects via In-Hand Pivoting using \\Vision, Force and Touch}
%\title{\LARGE \bf Reorienting Objects via In-Hand Pivoting using Vision, Force and Touch}
 
\author{Shiyu Xu$^{1}$, Tianyuan Liu$^{1}$, Michael Wong$^{1}$, Dana Kulić$^{1}$, Akansel Cosgun$^{2}$
\thanks{\hspace*{-1em}$^{1}$Monash University, Australia\newline
$^{2}$Deakin University, Australia}
\thanks{This work was supported in part by D. Kuli\'c's Australian Research Council Future Fellowship (FT200100761).}}
\maketitle

\begin{abstract}

%We propose a robotic manipulation system that can reorient objects laying on a surface by employing sensor-based control using vision, force and touch sensing.
%We study r
We propose a robotic manipulation method that can pivot objects on a surface using vision, wrist force and tactile sensing. We aim to control the rotation of an object around the grip point of a parallel gripper by allowing rotational slip, while maintaining a desired wrist force profile. Our approach runs an end-effector position controller and a gripper width controller concurrently in a closed loop. The position controller maintains a desired force using vision and wrist force. The gripper controller uses tactile sensing to keep the grip firm enough to prevent translational slip, but loose enough to allow rotational slip. Our sensor-based control approach relies on matching a desired force profile derived from object dimensions and weight, as well as vision-based monitoring of the object pose. The gripper controller uses tactile sensors to detect and prevent translational slip by tightening the grip when needed. Experimental results where the robot was tasked with rotating cuboid objects 90 degrees show that the multi-modal pivoting approach was able to rotate the objects without causing lift or translational slip, and was more energy-efficient compared to using a single sensor modality or pick-and-place.

\end{abstract}
\section{Introduction}
% \vspace{-0.01cm}

\begin{figure}[t!]
    \centering
    \includegraphics[trim={5 20 0 30}, clip, width=0.32\linewidth]{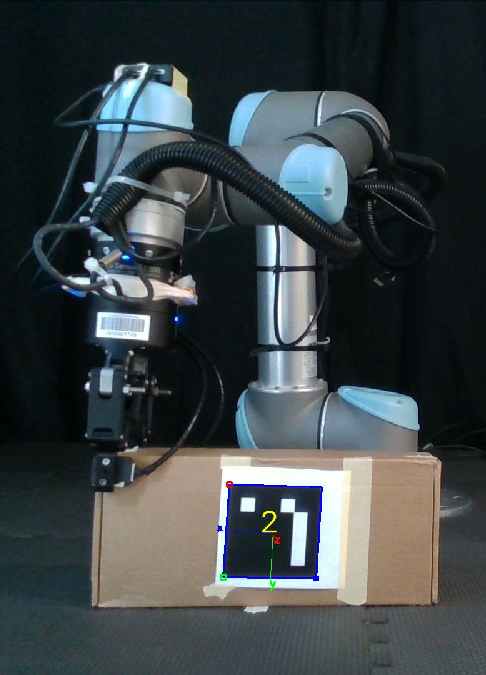} 
    \includegraphics[trim={10 0 0 0}, clip, width=0.32\linewidth]{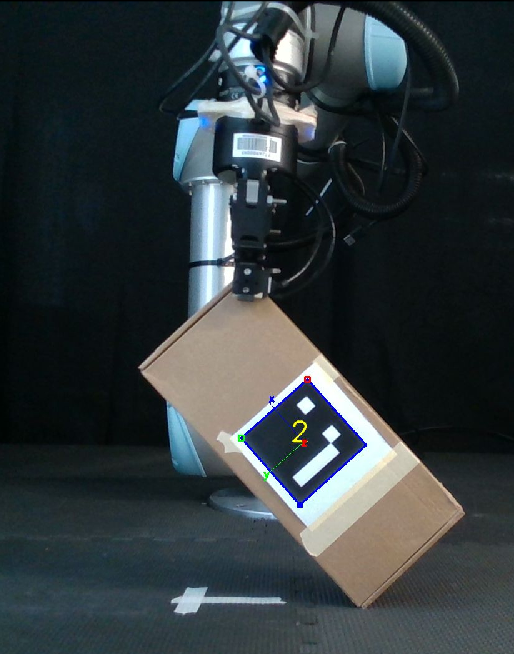}
    \includegraphics[trim={60 15 10 0}, clip, width=0.32\linewidth]{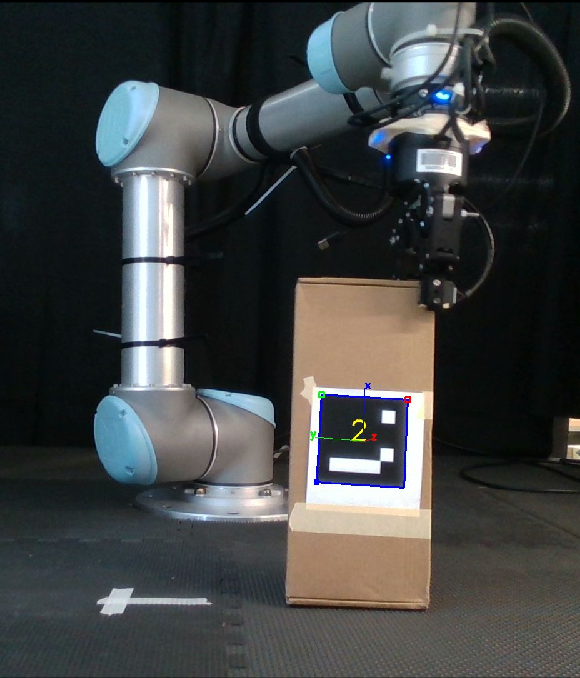}
    \caption{The robot pivots a box by 90 degrees in one motion while maintaining contact between the box and the surface. The end effector position is controlled using vision and wrist force sensing while the gripper width is controlled using tactile sensing.}
    \label{fig:intro}
\end{figure}

Robotic manipulation can be prehensile or non-prehensile. Prehensile manipulation involves capturing the object (e.g., via grasping) and requires achieving stable control over the grasped object. During object transport, prehensile manipulation requires the object to be lifted and held stably in the grasp. Non-prehensile manipulation, on the other hand, is a type of manipulation where objects are manipulated without grasping them. Non-prehensile manipulation allows a robot to perform a wider range of tasks than prehensile manipulation and can be more energy-efficient than prehensile manipulation because it does not require the robot to use as much force to move the object. In this study, we focus on performing pivoting actions on objects grasped by a parallel gripper. As shown in Fig.\ref{fig:intro}, this action takes an interesting middle ground between non-prehensile and prehensile manipulation. On the one hand, it requires the object to be grasped and gives the robot more stable control over the pose of the object. On the other hand, the action does not require the target object to be lifted, requiring much less effort than performing a full pick and place \cite{aiyama1993pivoting}. With less force exerted, this approach allows the manipulation of objects heavier than the maximum payload of a robot \cite{saeedvand2021hierarchical}.

% \begin{figure}[ht!]
%     \centering
%     %\def\figheight{5cm}
%     \begin{subfigure}[t]{0.49\columnwidth}
%     \centering
%     \includegraphics[trim={1500 400 300 250}, clip, width=\columnwidth,angle=270]{Pics/final pivoting 1.jpg}
%     \label{subfig: new pivoting1}
%     \end{subfigure}
%     %\hfill
%     %
%     \begin{subfigure}[t]{0.49\columnwidth}
%     \centering
%     \includegraphics[trim={1500 400 300 250}, clip, width=\columnwidth,angle=270]{Pics/final pivoting 2.jpg}
%     \label{subfig: new pivoting2}
%     \end{subfigure}
    
%     \caption{The robot pivots a box by 90 degrees. Our approach uses end-effector position control and gripper width control in a closed loop. Vision and wrist force are used to maintain a desired force profile using position control. A gripper-width controller, which is running concurrently, uses tactile sensing to keep the grip firm enough to prevent translational slip, but loose enough to induce rotational slip.}
%     \label{fig: new pivoting}
    
% \end{figure}

This study investigates reorienting an object grasped by a parallel gripper using pivoting actions. To achieve in-hand manipulation with a simple parallel gripper with no dexterous capabilities, we make use of extrinsic environmental factors \cite{dafle2014extrinsic}. Specifically, once the object is grasped, gravity is used to induce slip and rotate the object in-hand when lifting the gripper. Meanwhile, friction with the surface provides an opposing force to fully pivot the object. Thus, to allow this friction, the object must also be kept in contact with the surface during the motion. 

\begin{figure}[t!]
    \centering
    \includegraphics[trim={0 50 0 50}, clip, width=0.8\linewidth]{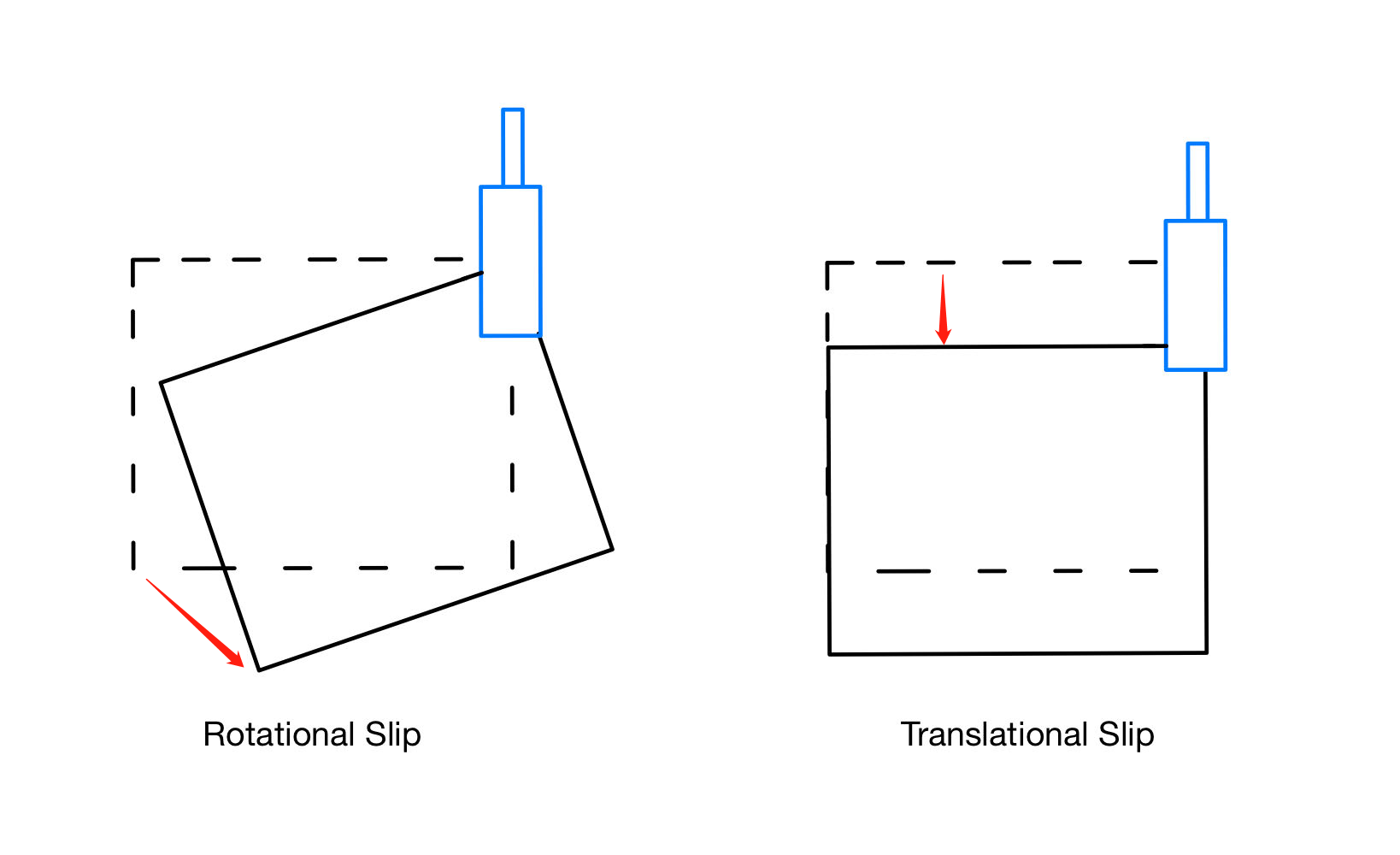}
    \caption{The two types of slip. The dotted outline describes the original position, and solid shape is the position after slip. The blue object is the gripper fingers.}
    \label{fig: slip types}
\end{figure}

As shown in Fig.\ref{fig:intro}, to avoid the need to turn the gripper along with the box, which introduces additional kinematic constraints on the robot, the box is allowed to rotate around the grasp point. Enabling slip in the grasp of the object is a non-trivial problem. To achieve the desired slip, the appropriate gripper force or finger width needs to be chosen in accordance to the properties of the object. To gain better insight into the forces that can induce in-hand slip, we follow other works on slip detection \cite{costanzo2018slipping, huh2020dynamically, meier2016tactile} in differentiating slip into two types, rotational slip and translational or linear slip. As shown in Fig.\ref{fig: slip types}, in rotational slip, the object is allowed to rotate while remaining grasped, with its centre of rotation at the grasp point and remaining in place as the object moves. Translational or linear slip involves the object moving away from the original grasping point. These two types of slip can occur on their own or simultaneously. For an in-hand pivoting task, the goal is to avoid translational slip, while allowing rotational slip \cite{toskov2022hand}.

Similarly, the trajectory of the pivoting motion also needs to be adjusted to ensure the object remains in contact with the surface. Visual information may be used to determine the object's pose, but contact with the surface may be challenging to estimate visually. On the other hand, force data can be used to derive whether the object is lifted by the robot, but otherwise would provide little information about the object's pose. 

We propose using multi-modal sensing in closed-loop control that can adjust both the robot arm trajectory and the gripper width to complete a pivoting action. The object state 

% page splitting shenanigans for geometry and margins
\restoregeometry

\noindent will be continuously detected through force/torque data at the parallel fingers and wrist of the gripper, as well as visual information from an RGB-D camera. Pivoting of the grasped object will then be achieved through motion planning with the arm's end-effector coupled with a position controller, while slip will be managed by adjusting the gripper width. 
Overall, the main contributions of this study are: 
\begin{itemize}
    %\item A slip detection algorithm using tactile force data as input, capable of differentiating translational and rotational slip
    \item A vision and force-based position controller for the arm trajectory that integrates an analytical force profile and object pose information
    \item A tactile sensor-based gripper width controller for parallel grippers that tightens based on detected slip
    %\item Experiments that validate both the slip detection algorithm and the vision and force-based position controller
    \item Experiments on a real robot that validate the effectiveness of multi-modal sensing for the pivoting task
    %\item A controller for the gripper finger width that utilises tactile sensor data
    %\item Evaluation metrics includes success rate, work consumption, slip off rate and lift up rate are used to validate the performance of our proposed approach.
\end{itemize}

\section{Related Works}

Pivoting provides an alternative to pick-and-place methods for reorienting the pose of an object. 
% Previous works involve both single arm \cite{toskov2022hand, antonova2017reinforcement, raessa2021planning} and dual arm \cite{zhang2021manipulation, shi2020aerial, yoshida2010pivoting} implementations to perform this task. 
A significant advantage of the pivoting motion is the ability to extend manipulation to reorienting heavy, large and/or long objects \cite{raessa2021planning, zhang2021manipulation, shi2020aerial, yoshida2010pivoting} that cannot be picked up by robot.  
Previous works focus on in-hand pivoting where the object is held in air \cite{toskov2022hand, antonova2017reinforcement} while others target pivoting using a surface to support the motion \cite{raessa2021planning, zhang2021manipulation, shi2020aerial, yoshida2010pivoting}. For our purposes, we will focus on pivoting with the assistance of a surface.

To achieve pivoting, planning based approaches are frequently adopted \cite{raessa2021planning, zhang2021manipulation, yoshida2010pivoting, hou2018fast, hou2019reorienting}. Closed loop control is also common, such as using impedance and admittance control to optimise a force planner \cite{shi2020aerial}. Machine learning methods are also used to control robot behaviour using complex input data \cite{toskov2022hand, antonova2017reinforcement}. Our solution aims at a learning-free control algorithm for the gripper width as well as end effector position during a pre-planned pivoting motion.

% Pivoting provides an alternative to pick-and-place methods for reorienting the pose of an object. Previous works involve both single arm \cite{toskov2022hand, antonova2017reinforcement, raessa2021planning} and dual arm \cite{zhang2021manipulation, shi2020aerial, yoshida2010pivoting} implementations to perform this task. Some of these works focus on in-hand pivoting where the object is suspended in air \cite{toskov2022hand, antonova2017reinforcement} while others target pivoting using a surface to support the motion \cite{raessa2021planning, zhang2021manipulation, shi2020aerial, yoshida2010pivoting}. For our purposes, we will focus on pivoting with the assistance of a surface. Many of these papers focus on reorienting heavy, large and/or long objects \cite{raessa2021planning, zhang2021manipulation, shi2020aerial, yoshida2010pivoting} with the intention of improving the dexterity of the robot, however we consider pivoting smaller and lighter objects that a robot can easily handle. As opposed to learning based methods \cite{toskov2022hand, antonova2017reinforcement} and path planning based approaches \cite{raessa2021planning, zhang2021manipulation, yoshida2010pivoting} our solution focuses on using a control algorithm to accomplish the set goal \cite{shi2020aerial}. However, the control algorithm we will develop is aimed at controlling the gripper width as well as position control from the set goal, whereas Shi \textit{et al.} concentrates on impedance and admittance control to optimise the force planner they designed for their aerial robot.

\subsection{Contact Control}

Adopting a purely open loop approach can often be limiting as the robot only considers the environment's initial states and is unaware how its interactions influence them \cite{siciliano1999robot}. In terms of the pivoting action, the motion of the robot needs to be updated based on the object's contact with the surface to induce friction and complete a full pivot. 

For similar tasks, force and torque data is often used to estimate the state of the robot held object's contact with its surroundings. Ma \textit{et al.} detected extrinsic contact between a grasped object and the environment using tactile sensors mounted to a parallel gripper \cite{ma2021extrinsic}. Molchanov \textit{et al.} used a data-driven approach, training machine learning models with tactile data to perform regression and classification for the presence and location of contact between object and environment \cite{molchanov2016contact}. Doshi \textit{et al.} used force/torque data to estimate how a wrench applied to an object would affect its motion using a contact model \cite{doshi2022manipulation}. Hogan \textit{et al.} used high-resolution tactile sensors to localise the pose of held object and estimate contact and slip status. Both \cite{doshi2022manipulation} and \cite{hogan2020tactile} also developed and experimentally tested closed loop controllers for manipulation.

Vision inputs are also used to compliment tactile data. Yu \textit{et al.} trained a model with visually detected pose estimates for an object of known geometry, as well as tactile data from a wrist force/torque sensor and robot encoders \cite{yu2018realtime}. The model was used to detect the contact arrangement of a held object, and achieved more accurate estimates than using vision alone. 

We take a multi-modal approach to maintain extrinsic contact, using vision-based pose estimation and force-based position control compared with a simpler vision-only method.
% 
% Adopting a purely position control approach can often be limiting as the robot is only provided with information on where it is, therefore, it is unaware of how it is interaction with the environment \cite{siciliano1999robot}. Robotic force control is especially useful in sensitive environments where disturbances or damage to the environment and/or the object that is being manipulated could be catastrophic. As a result, many works have researched the practicality of utilising a robotic force control in surgical settings \cite{xie2010force, zemiti2007mechatronic, davies1997active, federspil2003development}. Robots have the advantage of being more precise than humans are which is essential when performing surgery \cite{federspil2003development}. While our work does not involve a robot manipulating objects in a sensitive environment, force control does assist our research by preserving the objects we use, as well as reducing the risk of damaging the robot. Xie \textit{et al.} employs a hybrid position-force control architecture that starts with position control to move the needle till it initiates contact with the cell before switching to force control \cite{xie2010force}. Our approach adopts a similar methodology, which is detailed further in Section IV. However, our approach utilises vision sensors to control the robot as well. 

% \begin{figure*}[!ht]
%     \centering\includegraphics[trim={180 180 100 180}, clip, width=1\linewidth]{Pics/new_Module diagram.drawio-1-compressed.pdf}
%     \caption{System diagram }
%     \label{fig: module}
% \end{figure*}

\subsection{Slip Control}       % slip, estimation of translational/rotational
To achieve an angled pivoting position with relatively simple motions, gravity is used to induce slippage at the fingertips of the gripper, allowing change of pose without having to regrasp an object. To detect slip, tactile sensing at the contact with the object is often used \cite{li2020review}. Costanzo \textit{et al.}'s line of work investigated, separately, the prevention of both rotational and translational slip, as well as enabling rotation while preventing linear slip \cite{costanzo2018slipping, costanzo2019two, costanzo2023detecting}. They input force and torque data into an analytical friction model to estimate slippage, and develop a controller that prevents undesired slip with minimum force \cite{costanzo2023detecting}. Wang \textit{et al.} similarly used a friction model to develop a slip estimation algorithm and a gripper controller \cite{wang2021status}.
Data-based approaches are also prominent, with Toskov \textit{et al.} and Chen \textit{et al.} both training machine learning models to enable slip under gravity, with Toskov \textit{et al.} focusing on rotational slip \cite{toskov2022hand} and Chen \textit{et al.} focusing on translational slip \cite{chen2021tactile}.
While existing works predominantly pivot the objects in the air \cite{costanzo2023detecting, wang2021status, toskov2022hand, chen2021tactile}, we will attempt to detect and control slip while maintaining contact between the object and the surface it rests on, which allows the robot to reorient objects without needing to pick up and lift them, reducing effort required.

% \section{Problem Statement}
% \input{Statement.tex}

\section{Approach}
For simplicity, we focus on box-shaped objects with only one of three dimensions within the gripper's graspable width. 

\begin{figure}[ht!]
    \centering
    \includegraphics[width=1\linewidth]{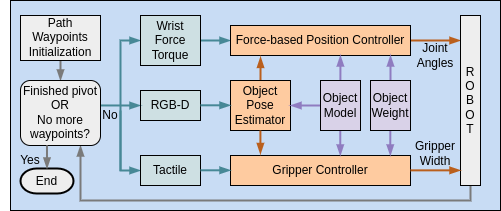}
    \caption{System diagram}
    \label{fig: module}
\end{figure}

The pivoting task is divided into three sub-tasks. The first task covers visual object pose detection where the 6D pose of the box is estimated. The second task generates the grasp pose, which is placed on a point along the top edge of the box based on a user input. Finally, closed loop gripper width control is used to grasp the object, and further regulates slip during the pivot motion. A force-based end-effector position controller is also used to amend a pre-planned end-effector path, aiming to maintain contact between the grasped box and the surface. The control loop runs until the robot is able to complete the pivoting task, or when all waypoints of the pre-planned path have been executed. This overall structure of the system is illustrated in Fig.\ref{fig: module}.

% \begin{figure*}[ht]
%     \centering\includegraphics[scale=0.28]{Pics/flowchart.png}
%     \caption{Flowchart of the system}
%     %the robot will generate a grasp pose from visual image captured by a camera, then try to pivot the object in an arc from its current pose up to a certain degrees either clockwise or anti-clockwise (depends on where the robot will grasp the object). The radius of the arc is determined from the object's dimensions.}
%     \label{fig: flowchart}
% \end{figure*}

\subsection{Object Pose Estimation}
\label{sec:box_pose_estimation}

An RGB-D side camera is used to detect the ArUco marker placed in the center of the largest surface of the box. Based on this marker detection as well as the box dimensions, the box pose can be inferred by offsetting the marker's pose by half of the box's respective dimension. The 6D pose of the box's axis of origin is defined at its centre point with an orientation that aligns the axes so that they are parallel with the length, width and height of the box. 

\subsection{Grasp Pose Synthesis} 

From the box's pose, a feasible grasp pose is generated using the known dimensions of the box. The gripper grasps the box such that the gripper is perpendicular to the table surface, i.e. the gripper is pointing down into the table. The grasp pose is also positioned ensuring that the sensing elements of the tactile sensors make full contact with the box. Given that there is only one graspable dimension, the robot can only grasp along one surface of the box. %The user is able to input a number to choose which end of the box should be grasped.  

The placement of the grasp pose on the top corner of the box prevents the box from being stopped by the gripper palm during the pivoting motion, and allows it to fit through the gap between the fingers. The grasp point relative to the box, as well as the orientation of the gripper, is maintained during the pivoting action. For a given grasp point, the pivot direction is always chosen such that the grasp point and pivot point are on opposite sides of the box. This is demonstrated in Fig.\ref{fig:intro}.

\subsection{Closed Loop Control}\label{closed_loop}
To successfully pivot the box we employ a robot end-effector position controller and a gripper controller. These controllers will adjust both the robot end-effector path and the gripper width respectively throughout the pivoting motion.

\begin{figure}[h]
        \centering\includegraphics[width=0.3\columnwidth]{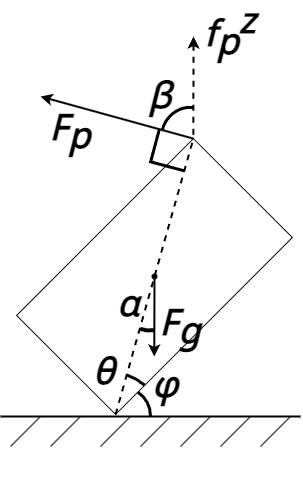}
        \caption{Free-body diagram of a pivoting box under equilibrium conditions}
        \label{fig: force_profile}
    \end{figure}

We analyse the expected forces applied to the sensors during the pivot process assuming equilibrium conditions during the pivot action, corresponding to a constant speed rotation, as illustrated in Fig.\ref{fig: force_profile}. Torque applied on the box by the gravitational force $F_g$, with the centre of mass assumed to be at the centre of the box, is balanced by the forces applied to it from the pivoting motion $F_p$. 
    
    The wrist sensor measured force in the vertical (z) direction is the vertical component of the pivoting force $F_p$, $f_p^z$:

    \begin{equation}
        f_p^z = \frac{F_g sin(\alpha)cos(\beta)}{2}
    \end{equation}

    The above equation can then be expressed with respect to the pivot angle between the base of the box and the surface, $\varphi$, and specific properties of the box, including the angle between its base and diagonal, $\theta$, its mass $m$, and gravity constant $g$:

    \begin{equation}\label{eq:pivot_force}
        f_p^z(\varphi) = \frac{mg sin(\frac{\pi}{2} - \varphi - \theta)cos(\varphi + \theta)}{2}
    \end{equation}

    This expected force profile is used in relation to the force measured by the wrist sensor to control the robot.

%\par\vspace{\baselineskip}
%\begin{enumerate}
    %\item 
    \noindent\textbf{Initial Gripper Width:} Determining an adequately loose grip is essential to allow the box to rotate in-hand. To grasp the box, the gripper fingers are set to continue closing until the total force in the direction normal to the box surface, as measured by the two tactile sensors, exceeds a threshold. 
    
    This force threshold is set to ensure that the gripper does not experience translational slip initially. This requires the static friction force between the gripper fingers and the box to be greater than the pivoting force $f^z_p$ applied to the box at the start of the motion, when the pivot angle $\varphi$ is infinitesimally small. The static friction force $F\textsubscript{static}$ is calculated from empirically tested friction coefficient $\mu_s$ and the normal force $F_N$ exerted through contact as measured by the tactile sensors, giving the grip force threshold: 

    \begin{equation}
        F\textsubscript{static} = \mu_sF_N  > f_p^z(\varphi \approx 0)
    \end{equation}
    
    % \begin{equation}
    %     F_N > f_p^z(\varphi \approx 0) % init_ft_force from code
    % \end{equation}
    
    % Additionally, to speed up the grasping process, the gripper first closes to a distance slightly larger than the known dimensions of the box before executing the grasping algorithm. The fingers do not make contact with the box during this action.

    \noindent\textbf{Force-based Position Controller:} The robot force-based position controller controls and updates the arm's path throughout the pivoting motion. % This controller utilises the force/torque wrist sensor which provides force readings, and the RGB side camera which detects and estimates the box's rotation angle as the robot pivots the box. 
    
    We represent a complete pivoting motion with the pivot point of the object remaining in contact with the surface by the ideal pivoting force profile described by Equation \ref{eq:pivot_force}. The controller will then maintain contact by tracking this profile. The force exerted by the robot is controlled by applying a vertical offset to a pre-planned arc path for the end-effector.

\begin{figure}[!ht]
        \centering\includegraphics[width=0.7\linewidth]{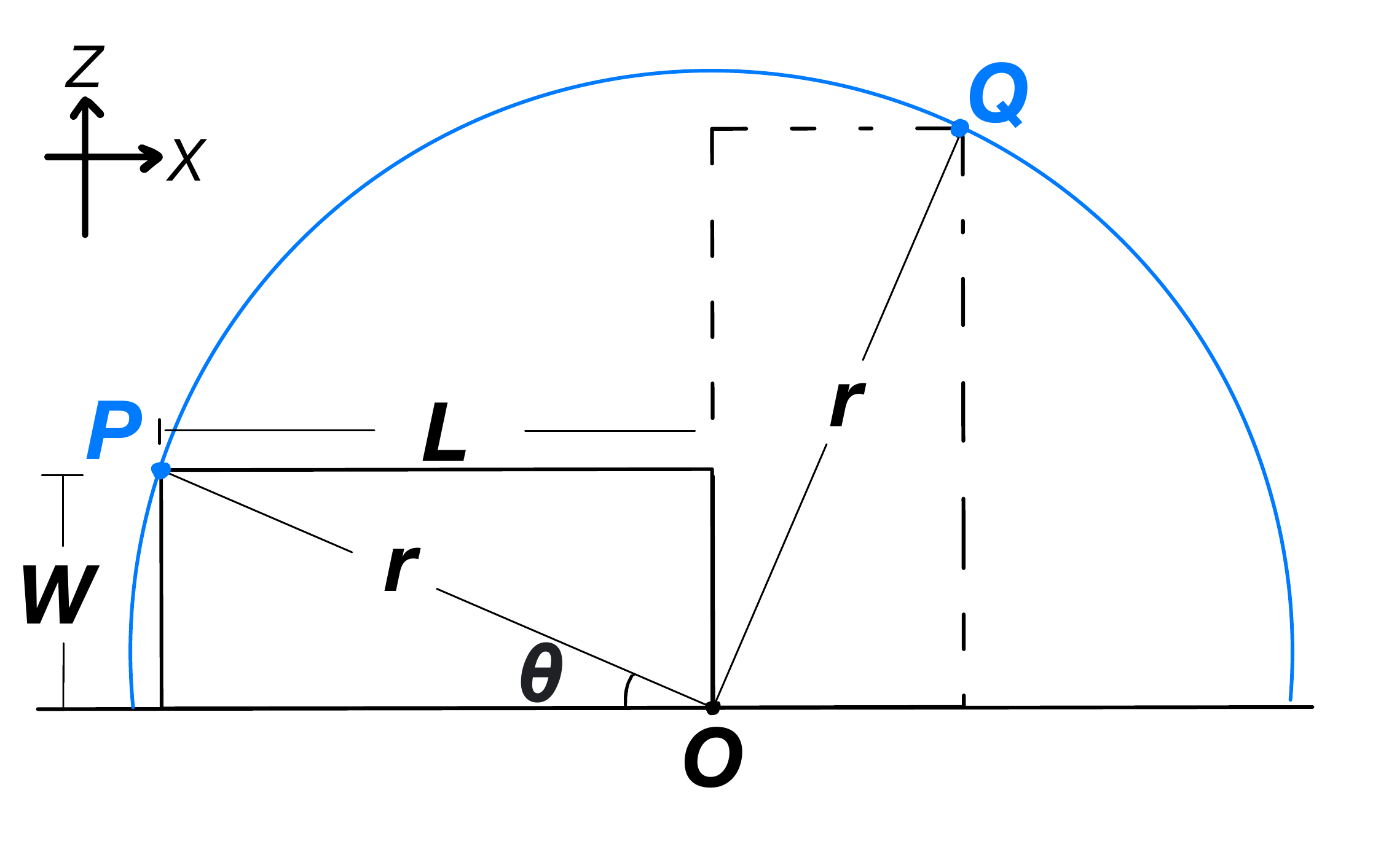}
        \caption{Path Analysis: the box is pivoting from the solid line position to dashed line position. O represents the pivoting center, P is the grasp point at the beginning of the pivot and Q is the corresponding point of P after pivoting. When the grasp point is at point P, the corresponding pivot angle $\varphi$ is 0\degree, whereas at point Q $\varphi$ is 90\degree.}
        \label{fig:arc}
    \end{figure}

    Initially, a Cartesian path consisting of 50 waypoints is generated to traverse an arc parameterised by the box's dimensions. 50 waypoints are used as a balance between time taken to complete the movement and maintaining the arc with radius as the distance between the grasp point and the pivot point. For a box grasped at the top corner, the radius would become its diagonal, as seen in Fig.\ref{fig:arc}. 

    To pivot the box from point $P$ to point $Q$, the movement required to get to each waypoint on the arc trajectory from the starting point $P$ is given as:
    \begin{equation}\label{eq:path_x}
        \Delta x = L - r \cdot cos(\varphi + \theta)
    \end{equation}
    \begin{equation}\label{eq:path_z}
        \Delta z = r \cdot sin(\varphi + \theta) - W
    \end{equation}

    where $L$ is the length of the box, the arc radius $r$ is the diagonal of the box and $\varphi$ is the pivot angle between the surface and the base of the box.

    The MoveIt motion planning API is then used to generate and execute the timed trajectory for the robot end effector between each waypoint. It computes straight line path segments between the start and end using linear interpolation. The velocity and acceleration of the trajectory is then derived with an algorithm termed iterative parabolic time parameterisation, which generates parabolic blends to smooth the planned paths and adhere to velocity and acceleration limits \cite{kunz2011turning}. 
    % The robot will move according to this pre-generated plan, while the controller continuously checks that the box's rotation angle does not exceed the goal rotation angle, which is 90 degrees for a complete pivot, to ensure the box stops appropriately.  

    % To close the loop for dynamic pivot action, we use an analytical force profile in the vertical dimension to detect and ensure that the box remains in contact with the surface. 
    % From empirical tests, the relationship between the height that the gripper lifts the box up to and the force measured by the wrist sensor during surface-contact pivoting was found to be roughly linear. Thus, to control the force applied to the box, a vertical offset is applied to the trajectory for every waypoint. 

    Using the above information including Equation \ref{eq:pivot_force}, the position controller is implemented, and described in Alg.\ref{alg:traj}. At each waypoint, the ideal force $f_{ideal}$ is predicted using the rotation angle of the box, estimated using vision. The measured force $f_{real}$ is then compared to the ideal force to produce an error, and accumulated in $error_{accum}$. The vertical offset that should be applied is then calculated from the force error using Proportional-Integral (PI) control, with proportional constant $K_p$ and integral constant $K_i$, both empirically tuned. The offset is applied to all waypoints, and adjusted incrementally during each iteration. This aims to maintain the offset that minimises the error between the two force values. The pre-planned waypoint is updated by applying this offset to the vertical dimension, and trajectories are generated by MoveIt so that the robot moves in accordance to the adjustment. If at any point the rotation angle is estimated to be greater than or equal to 90\degree, it is assumed that the pivot is complete and the program will terminate. 

    \begin{algorithm}
    \caption{Force-based position controller, using z as the vertical dimension}\label{alg:traj}
    \begin{algorithmic}
    \State $\mathit{offset} \gets 0$ 
    \State $error_{accum} \gets 0$
    \FOR{\textit{waypoint} \textbf{in} \textit{path}}
        \State $\varphi \gets $ vision detection
        \State $f_{real} \gets $ force torque sensor
        \State $f_{ideal} \gets $ $f_p^z(\varphi)$ (Equation \ref{eq:pivot_force})
        \IF{$\varphi < 90\degree$}
            \State $error \gets |f_{ideal} - f_{real}|$
            \State $error_{accum} \gets error_{accum} + error$
            \State $\mathit{offset} \gets \mathit{offset} + K_p \times error + K_i \times error_{accum}$
            \State $waypoint.z \gets waypoint.z + \mathit{offset}$
            \State move\_robot($waypoint$)
        \ELSE
            \State \textbf{break}
        \ENDIF
    \ENDFOR
    \end{algorithmic}
    \end{algorithm}

    Due to the MoveIt planning computation, the robot pauses momentarily before each consecutive movement to the next waypoint. This causes a discontinuous, slow motion that becomes slower with more waypoints. 

    \textbf{Gripper Controller:} The slip of the object is controlled to perform the pivoting movement. We estimate the slip type as either translational or rotational. During pivoting, the desired slip is rotational slip and translational slip as a result of gravity should be avoided. We ignore the no slip case as this was not possible for our chosen configuration. Hence it is sufficient to only explicitly detect and control translational slip.
    
    To detect the type of slip, we utilise the direction of force measurements at different points of the tactile sensing elements. Our method uses distributed sensing elements on the tactile sensor, and assumes the tactile sensor consists of nine physical elements in a 3x3 square array \cite{khamis2018papillarray}. Each element measures the displacement of the tip of the sensing pillar, and predicts applied force from this displacement. We focus on the displacement and force values in the vertical dimension, parallel to gravity, as this is the major indicator of slip. 
    
    In the event of gravity-induced translational slip, the downwards movement of the box will be captured by the tactile sensor as a global downwards force across all sensing elements on the tactile sensor. This is in contrast to when the box undergoes rotational slip, where local sensing elements measure forces in different directions, as the centre of rotation is roughly at the centre of the sensor. The direction of the applied force in the vertical dimension is simply taken as the sign of the measured value. We also set a small magnitude threshold of 0.1 to exclude noise. Slip type is only estimated for sensing elements that are detected to be in contact, through an in-built algorithm of the tactile sensors which is based on the normal force applied to the element. 
    
    If the tactile sensors detect translational slip, the gripper width incrementally closes until it is no longer detected. Each step of the increment is $1/256$ times the max distance between the gripper fingers. If the tactile sensors detect rotational slip, then the gripper width is unaltered. The gripper will also loosen its grip if at least one of the pillars experience excessive deformation, to prevent damage to the sensors and the box. We choose the displacement threshold to be slightly higher than values seen in trial runs keeping the heaviest box in midair while avoiding slip, representing the expected highest force situation. The gripper control algorithm is summarised in Alg. \ref{alg:grip}. It is executed at the tactile sensor sample rate of 500 Hz.

    \begin{algorithm}
    \caption{Gripper controller}\label{alg:grip}
    \begin{algorithmic}
    \State $downwards\_z\_count \gets 0$
    \State $contact\_count \gets 0$
    \FOR{\textit{element} \textbf{in} \textit{sensor array}}
        \IF{element detects contact}
            \State $contact\_count \gets contact\_count + 1$
        \ENDIF
        \IF{$displacement_z < -0.1$}
            \State $downwards\_z\_count \gets downwards\_z\_count + 1$
        \ENDIF
        \IF{$displacement_{x,y,z} > 5$}
            \State loosen\_gripper()
        \ENDIF
    \ENDFOR
    \IF{$downwards\_z\_count == contact\_count$}
        \State tighten\_gripper()
    \ENDIF
    \end{algorithmic}
    \end{algorithm}
%\end{enumerate}

\section{Experiments}

\subsection{Setup}
\begin{figure}[t]
  \centering   
  \begin{overpic}[trim={0 520 0 0}, clip, width=0.96\linewidth]{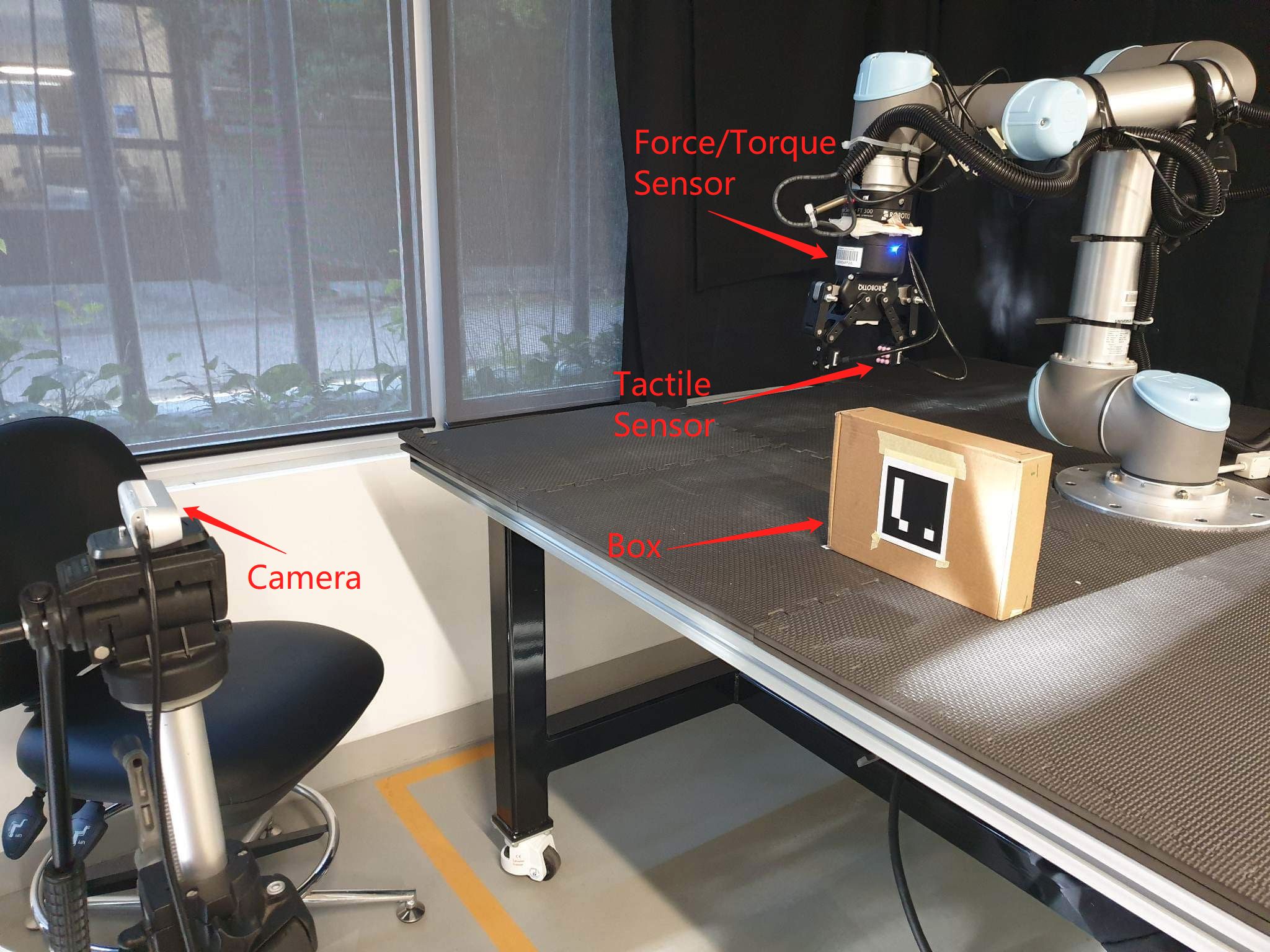}
     \put(2,16){\includegraphics[trim={220 300 200 0}, clip, frame, cfbox=white,  width=0.4\linewidth]{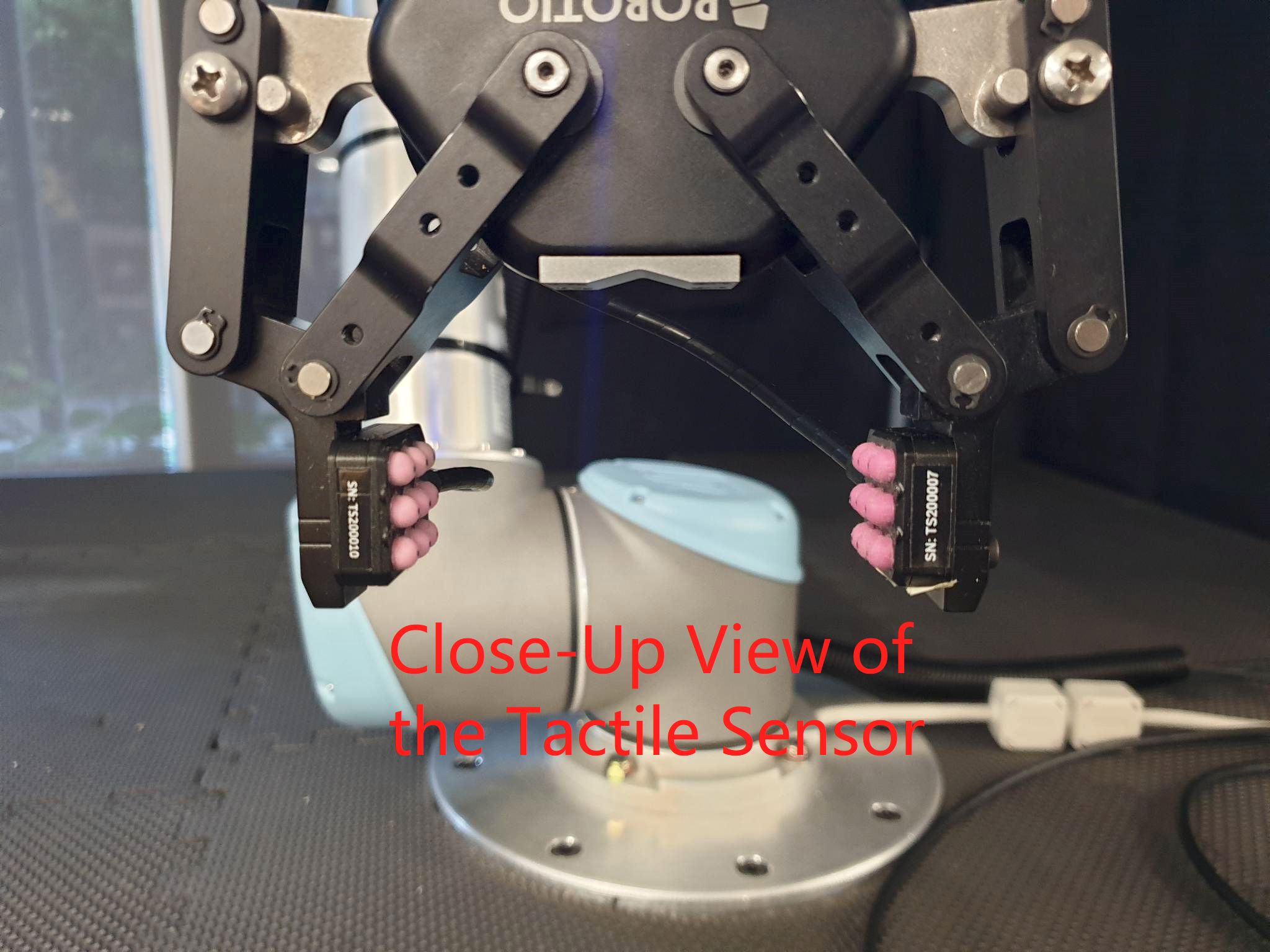}}  
  \end{overpic}
\caption{Experimental Setup}
\label{fig: set up}
\end{figure}

The proposed pivoting system was tested using a UR5 robot arm with the Robotiq 2F-85 gripper and a FT-300 Force/Torque sensor attached to the end effector. The fingers of the gripper have two Contactile PapillArray 3x3 tactile sensor arrays attached, capable of measuring 3D displacement, force, and torque \cite{khamis2018papillarray}. In some experiments involving high-force grasping, the tactile sensors are replaced with flat rubberised fingers to avoid damaging the sensors. The gripper has the tactile sensors mounted unless specified otherwise. The camera used is the Intel RealSense D435i, placed to the side of the robot. The hardware setup is shown in Fig. \ref{fig: set up}.

\begin{figure}[t]
    \centering\includegraphics[trim={0 450 0 450}, clip, width=1\linewidth]{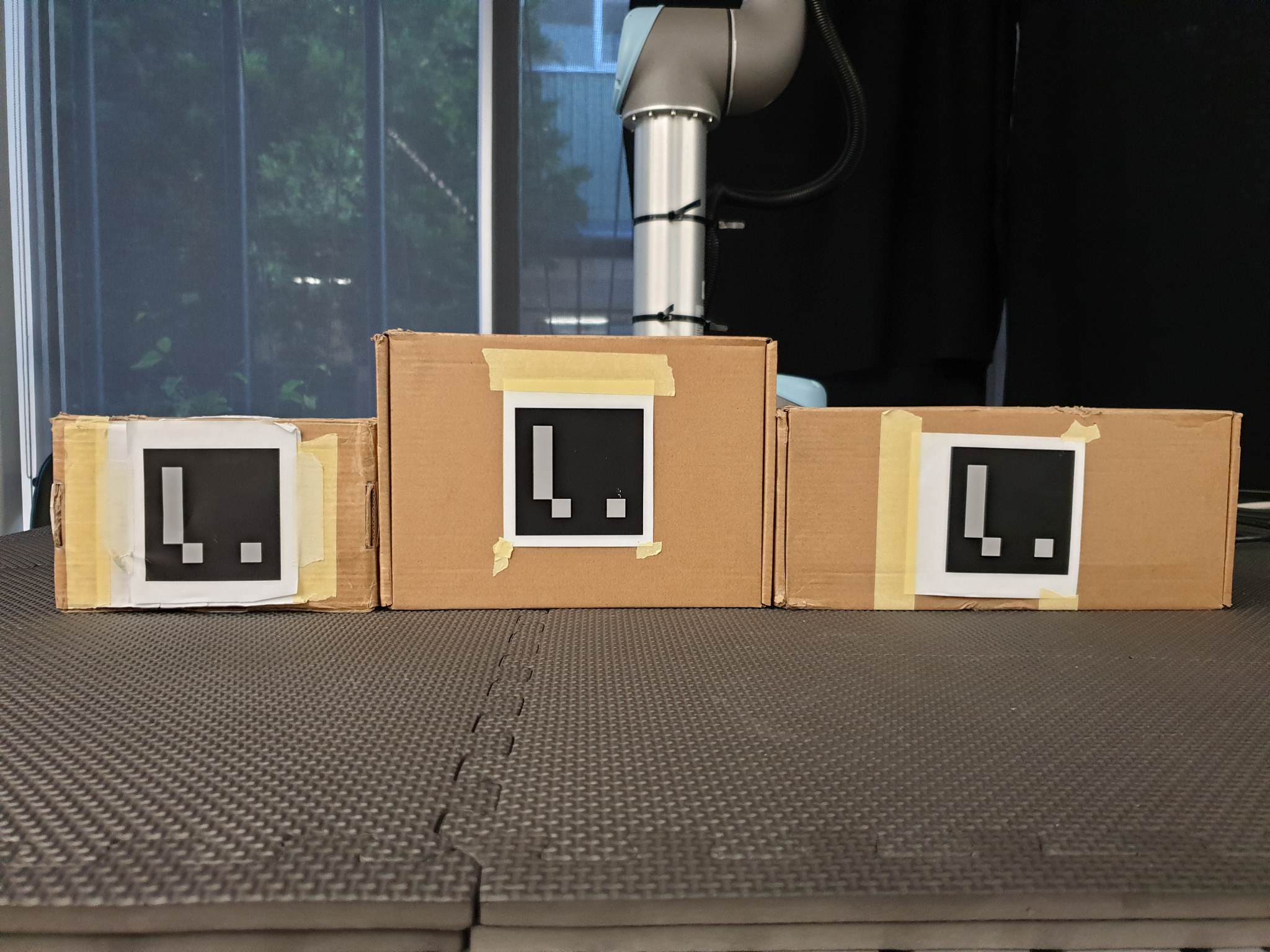}
    \caption{The cardboard boxes used for pivoting. From left to right: Small box (18x11x4cm, 1.27kg), Large box (23x16x5cm, 0.88kg), Long box (28x12x5cm, 1.72kg)}
    \label{fig: boxes}
\end{figure}

Cardboard boxes with varying dimensions are used for the experiment, shown in Fig. \ref{fig: boxes}. Their mass and mass distributions are adjustable by adding iron weights inside.

% \begin{figure}[t!]
%     \centering\includegraphics[trim={0 0 0 0}, clip, width=0.485\linewidth]{Pics/Scene.jpeg}
%     \includegraphics[trim={0 0 0 0}, clip, width=0.485\linewidth]{Pics/Scene.jpeg}
%     \caption{Experimental Setup}
%     \label{fig: set up}
% \end{figure}

\subsection{Procedure} 

A series of experiments are designed to verify the validity of our approach.  
The goal of all experiments is to complete a $90\degree$ rotation of the box without lifting it off the surface or letting the box slip out of the gripper's grasp. A simple pick-and-place method is  performed as a baseline. We compare five different methods, including the application of controllers introduced in Section \ref{closed_loop}, under four different conditions. Each applicable combination was repeated 10 times for each of the three boxes.

% In the baseline approach, the robot first lifts up the object completely and then rotates $90 \degree$ of the front joint, and completes pivoting after lowering it down. The gripper will be set to close as hard as possible for cases that don't have the gripper controller. 

\subsection{Methods}

\noindent\textbf{Pick\&Place:} Picking up a box from the centre of its top edge, the robot first lifts up the box completely and then rotates it in the air by $90\degree$, and completes pivoting after lowering it down. The gripper is set to close as hard as possible, with the tactile sensors replaced with flat, rubberised fingers. The lifted height before pivoting is set just high enough so that the robot can complete pivoting without hitting the surface, to represent the lowest effort situation using this method.

\noindent\textbf{Open Loop:} No controllers are used. The gripper is set to close as hard as possible and the tactile sensors are also replaced with flat, rubberised fingers. The arm's trajectory is planned as an ideal arc based on the box's dimensions.
 
\noindent\textbf{Vision:} A vision-based controller is used. The gripper is set to close as hard as possible with the tactile sensors replaced with flat, rubberised fingers. The arm's trajectory is planned as an ideal arc based on the box's dimensions and adjusted with offsets. However, the vertical offset is calculated from the estimated height of the box from the surface. 

\noindent\textbf{Gripper:} Only the gripper controller is used, both to grasp the object and to control slip during pivoting. The arm's trajectory is planned as an ideal arc, not updated during the movement.

\noindent\textbf{Force:} Only the force-based position controller is used. The gripper is set to close as hard as possible with the tactile sensors replaced with flat, rubberised fingers. The arm's path is planned as an ideal arc and updated during the movement using the error between measured force and predicted force. 

\noindent\textbf{Force+Gripper+Vision:} All three sensors are used. The gripper controller is used to grasp the object and control slip. Vision is used to track the rotation angle between the surface and the box's lowest edge for force estimation. The arm's path is planned as an ideal arc and updated during the movement by the force-based position controller, and the gripper width is updated by gripper controller.

% \subsection{Conditions}
% \begin{enumerate}
%     \item \textbf{Long-to-Short Pivot:} The robot pivots a box from standing on its longer edge to standing on its shorter edge. 
%     % The box will start on its longest graspable surface. The arm's trajectory planned is ideal. 
%     \item \textbf{Long-to-Short Pivot with added noise:} Using the same pivoting method, the robot will pivot the box from standing on its longer edge. Noise is added to the initially planned arc trajectory by adding 5cm to the base dimension of the box.
%     % However, the arm's trajectory planned is not ideal, i.e. parameters will be fed into the arc trajectory planning that will offset it from an ideal trajectory. 
%     \item \textbf{Short-to-Long Pivot:} The robot pivots a box from standing on its shorter edge to standing on its longer edge.
%     %The arm's trajectory planned is ideal.
%     \item\textbf{Short-to-Long Pivot with added noise:} Using the same pivoting method, the robot will pivot the box from standing on its shorter edge. Noise is added to the initially planned arc trajectory by adding 5cm to the base dimension of the box.
% \end{enumerate}

\subsection{Independent Variables}
\begin{enumerate}
    \item \textbf{Pivoting types:} The robot performs two types of pivoting. It either pivots a box from standing on its longer edge to standing on its shorter edge, or from its shorter edge to its longer edge. 
    \item \textbf{Noise:} A 5cm noise value for the base dimension of the box can be added, causing the arc trajectory to have a larger radius. No noise represents the ideal arc trajectory.
\end{enumerate}

\subsection{Evaluation Metrics} 

To assess the performance of the robot in each experiment, five quantitative measures are established:

\begin{itemize}
    \item \textbf{Success Rate (\%)}: The percentage of all attempts where the robot fully completes the $90 \degree$ rotation.
    \item \textbf{Slip Off Rate (\%)}: The percentage of all attempts where the box slips off from the gripper during the movement.
    \item \textbf{Lift Up Rate (\%)}: The percentage of all attempts where the box is lifted off the surface during the movement, regardless of whether contact was re-established later during the motion.
    \item \textbf{Time Taken (s)}: The total time it takes the robot to move the object from the start to end position.
    \item \textbf{Work (Joules)}: The amount of both translational work and rotational work done applied by the robot on the object to move the object from the start to goal pose. 
    
    % Work is defined as the energy transferred to or from an object via the application of force along a displacement. The measurements required to determine work done will be retrieved from the PapillArray tactile sensors and the FT-300 Force/Torque sensor. This criteria will demonstrate how much effort is required of the robot to complete the task.
\end{itemize}

Slip off and lift up events are observed by the experimenter, while the work is calculated from force measurements from the wrist mounted force/torque sensor, and displacement derived from the object pose estimation. 

A pivot action is considered failed when the robot cannot complete the rotation. Lift up and slip off rates are used to distinguish between perfect success (i.e. neither leaving nor being pushed too hard into the surface) and conditional success, which is less efficient in terms of effort and has a higher risk of damaging the object or environment.

\subsection{Results}
The detailed results for all experiments can be found in Table \ref{table:box_1_results}, with the aggregate results presented in Table \ref{table:all_results}. The combined controller approach achieves 100\% rate with 0 lift up and 0 slip off events. This approach also requires the least work among all approaches while the pick\&place method required the highest work. However, it lags behind other methods in terms of completion time. Instead, the open loop approach requires the least time to complete the movement.

\begin{table*}[htbp]
\centering
\caption{Experiment results summary.`NA' indicates no such cases were tested since no dimensions of the boxes are used in the pick\&place method. `-' is used for cases where the robot failed to do pivoting in all experiments.}
\begin{tabular}{|l|ccccc|ccccc||ccccc|ccccc|}
\hline

& \multicolumn{10}{c||}{Short-to-Long Pivot} & \multicolumn{10}{c|}{Long-to-Short Pivot}\\ \hline
& \multicolumn{5}{c|}{Without Noise} & \multicolumn{5}{c||}{Added Noise} & \multicolumn{5}{c|}{Without Noise} & \multicolumn{5}{c|}{Added Noise}\\ \hline

& $\%$ & $\%$ & $\%$ & sec & J & $\%$ & $\%$ & $\%$ & sec & J & $\%$ & $\%$ & $\%$ & sec & J & $\%$ & $\%$ & $\%$ & sec & J\\ \hline
\rowcolor{LightCyan}
& Succ. & Lift & Slip & Time & Work & Succ. & Lift & Slip & Time & Work & Succ. & Lift & Slip & Time & Work & Succ. & Lift & Slip & Time & Work\\ \hline

\multicolumn{21}{|c|}{Small Box (Dimensions: $18 \times 11 \times 4 $ cm, Weight: $1.27$ kg) }\\ \hline
Pick\&Place           & 100     & 100 & 0 & \textbf{16.4} & 8.0 & NA & NA & NA & NA & NA & 100 & 100 & 0 & 18.2 & 8.0 & NA & NA & NA & NA & NA\\ \hline 
\rowcolor{LightGray}
Open Loop           & 0     & 100 & 0 & - & - & 0 & 100 & 0 & - & - & 100 & 0 & 0 & \textbf{10.4} & 2.7 & 0 & 100 & 0 & - & -\\ \hline
Vision            & 100     & 100 & 0 & 22.7 & 4.4 & 100 & 100 & 0 & \textbf{23.2} & 4.7 & 100 & 100 & 0 & 24.1 & 2.2 & 100 & 100 & 0 & 24.3 & 2.3\\ \hline
\rowcolor{LightGray} 
Gripper           & 100     & 0 & 0 & 27.0 & 1.6 & 0 & 100 & 0 & - & - & 100 & 0 & 0 & 29.6 & 1.9 & 0 & 90 & 10 & - & -\\ \hline
Force            & 100     & 100 & 0 & 22.9 & 6.4 & 100 & 100 & 0 & 23.5 & 5.8 & 100 & 0 & 0 & 23.8 & 3.1 & 100 & 80 & 0 & \textbf{21.7} & 2.1\\ \hline
\rowcolor{LightGray}
Gripper+Force+Vision           & \textbf{100}     & \textbf{0} & \textbf{0} & 25.3 & \textbf{1.1} & \textbf{100} & \textbf{0} & \textbf{0} & 24.4 & \textbf{1.7} & \textbf{100} & \textbf{0} & \textbf{0} & 26.4 & \textbf{1.5} & \textbf{100} & \textbf{0} & \textbf{0} & 24.9 & \textbf{1.8}\\ \hline

\hline
\multicolumn{21}{|c|}{Large Box (Dimensions: $23 \times 16 \times 5 $ cm, Weight: $0.88$ kg) }\\ \hline
Pick\&Place           & 100     & 100 & 0 & \textbf{16.1} & 6.6 & NA & NA & NA & NA & NA & 100 & 100 & 0 & 16.6 & 5.9 & NA & NA & NA & NA & NA\\ \hline 
\rowcolor{LightGray}
Open Loop           & 0     & 100 & 0 & - & - & 0 & 100 & 0 & - & - & 70 & 0 & 0 & \textbf{10.8} & 4.0 & 0 & 100 & 0 & - & -\\ \hline
Vision            & 0     & 100 & 0 & - & - & 0 & 100 & 0 & - & - & 100 & 90 & 0 & 24.4 & 5.5 & 100 & 100 & 0 & 28.1 & 3.1 \\ \hline
\rowcolor{LightGray} 
Gripper           & 100     & 90 & 0 & 26.8 & 2.4 & 30 & 100 & 10 & 28.3 & 4.3 & 100 & 0 & 0 & 26.7 & 1.9 & 100 & 100 & 0 & 34.0 & 2.5\\ \hline
Force            & 0     & 100 & 0 & - & - & 0 & 100 & 0 & - & - & 100 & 0 & 0 & 24.4 & 5.4 & 100 & 80 & 0 & \textbf{25.1} & 3.3\\ \hline
\rowcolor{LightGray}
Gripper+Force+Vision           & \textbf{100}     & \textbf{0} & \textbf{0} & 25.5 & \textbf{1.8} & \textbf{100} & \textbf{0} & \textbf{0} & \textbf{25.8} & \textbf{1.9} & \textbf{100} & \textbf{0} & \textbf{0} & 26.2 & \textbf{2.3} & \textbf{100} & \textbf{0} & \textbf{0} & 26.9 & \textbf{1.9}\\ \hline

\hline
\multicolumn{21}{|c|}{Long Box (Dimensions: $28 \times 12 \times 5 $ cm, Weight: $1.72$ kg)}\\ \hline
Pick\&Place           & 0     & 100 & 0 & - & - & NA & NA & NA & NA & NA & 0 & 100 & 0 & - & - & NA & NA & NA & NA & NA\\ \hline 
\rowcolor{LightGray}
Open Loop           & 90  & 10 & 0 & \textbf{9.3} & 8.7 & 0 & 100 & 0 & - & - & 100 & 0 & 0 & \textbf{10.8} & 12.7 & 0 & 100 & 0 & - & - \\ \hline
Vision            & 100     & 100 & 0 & 24.6 & 4.1 & 100 & 100 & 0 & 28.2 & 4.0 & 100 & 100 & 0 & 28.7 & 5.3 & 100 & 100 & 0 & 31.0 & 5.2\\ \hline
\rowcolor{LightGray} 
Gripper           & 90 & 0 & 10 & 28.7 & 3.6 & 90 & 100 & 10 & 35.1 & 9.8 & 100 & 0 & 0 & 29.8 & 7.0 & 0 & 100 & 50 & - & -\\ \hline
Force            & 100 & 0 & 0 & 23.9 & 4.5 & 100 & 100 & 0 & \textbf{24.9} & 4.9 & 100 & 0 & 0 & 26.1 & 7.1 & 100 & 0 & 0 & \textbf{27.4} & 4.8\\ \hline
\rowcolor{LightGray}
Gripper+Force+Vision           & \textbf{100}     & \textbf{0} & \textbf{0} & 29.3 & \textbf{2.9} & \textbf{100} & \textbf{0} & \textbf{0} & 30.4 & \textbf{3.3} & \textbf{100} & \textbf{0} & \textbf{0} & 30.6 & \textbf{3.9} & \textbf{100} & \textbf{0} & \textbf{0} & 33.1 & \textbf{3.7}\\ \hline

\end{tabular}
\label{table:box_1_results}
\vspace{-.3cm}
\end{table*}

\begin{table}[htbp]
\center
\caption{Aggregate results from 60 trials for the Pick\&Place method (Pick\&Place was not run for the noisy dimension condition), and 120 for each of the other methods.}
\begin{tabular}{|l|ccccc|}
\hline
& $\%$ & $\%$ & $\%$ & sec & J\\ \hline
\rowcolor{LightCyan}
& Succ. & Lift & Slip & Time & Work\\ \hline
Pick$\&$Place           & 66.7  & 100 & 0 & 16.8 & 7.1\\ \hline 
\rowcolor{LightGray}
Open Loop & 30  & 67.5 & 0 & \textbf{10.3} & 7.0\\ \hline 
Vision           & 83.3  & 99.2 & 0 & 25.9 & 4.1\\ \hline 
\rowcolor{LightGray}
Gripper           & 67.5  & 56.7 & 7.5 & 29.6 & 3.9\\ \hline 
Force           & 83.3  & 55 & 0 & 24.4 & 4.7\\ \hline 
\rowcolor{LightGray}
Force + Gripper + Vision          & \textbf{100}  & \textbf{0} & \textbf{0} & 27.4 & \textbf{2.3}\\ \hline 
\end{tabular}
\label{table:all_results}
\end{table}

\section{Discussion}
\subsection{Highest Performance and Robustness}
\textbf{Performance:} The combined approach of vision, force-based position and gripper control had the highest success rate and lowest work among all approaches. It was able to adjust both the gripper width and the movement path so that the object was not lifted off the ground. However, the long completion time of 27.4 seconds on average is the disadvantage of our approach. This is due to the motion planning of each subsequent end-effector waypoint when applying the control offset, which requires more time than other approaches. The fastest approach was the open loop method with an average completion time of 10.3 seconds, as it only plans once for the entire trajectory. On the other hand, this also indicates that the pivoting action is capable of faster completion than the traditional pick-and-place method (16.8 seconds on average). If a more responsive control methodology such as pure joint velocity control without planning was implemented, the closed loop solution may also achieve speeds approaching the open loop method. 

The lift and slip rates indicate the importance of the gripper controller. For methods that do not use gripper controller, closing the gripper as hard as possible avoids translational slip. This is reflected in the very low slip rate across all methods, even when the boxes have been lifted off the surface. Meanwhile, without the gripper controller, rotational slip could not be achieved consistently, and was prone to influences by other factors such as the box's mass, shape and mass distribution. Without rotating in the gripper, the pivot point would lose contact with the surface as the box cannot maintain a pose that keeps its diagonal in line with the radius of the trajectory. This corresponds to higher lift rates compared to methods with gripper controller.

\textbf{Robustness:} To test robustness, noise was intentionally added into the system in the form of a 5cm offset to one of the box dimensions. This had the effect of increasing the radius of the initial planned arc trajectory. As expected, the open loop approach always failed in this case. The single-modal closed loop approaches were still capable of completing full pivots in many cases, with success rates of 67.5-83.3\%, despite high lift rates of 55-99.2\%, corresponding to a frequent deviation from the ideal planned trajectory, but a recovery to complete the pivot. Out of all methods, the combined approach was the most robust, succeeding in all tests. 

\begin{figure}[t]
    \centering
    \begin{subfigure}{0.49\linewidth}
        \centering
        \includegraphics[width=\textwidth]{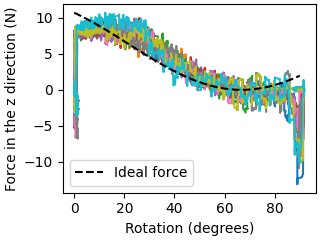}
        \caption{Force, without noise}
        \label{fig:force_plot_without}
    \end{subfigure}
    \hfill
    \begin{subfigure}{0.49\linewidth}
        \centering
        \includegraphics[width=\textwidth]{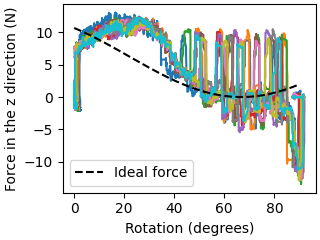}
        \caption{Force, with noise}
        \label{fig:force_plot_with}
    \end{subfigure}
%     \caption{The position in the z direction vs. the position in the x direction. Data from 10 trials are shown with each trial in a different colour, in addition to the ideal force profile specified in Equation \ref{eq:pivot_force}. The setup uses the long box performing long-to-short pivots using combined modalities.}
%     \label{fig:force_plot}
% \end{figure}

% \begin{figure}[t!]
%     \centering
    \begin{subfigure}{0.49\linewidth}
        \centering
        \includegraphics[width=\textwidth]{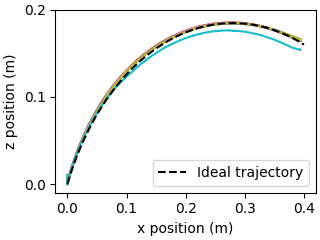}
        \caption{Trajectory, without noise}
        \label{fig:pos_plot_without}
    \end{subfigure}
    \hfill
    \begin{subfigure}{0.49\linewidth}
        \centering
        \includegraphics[width=\textwidth]{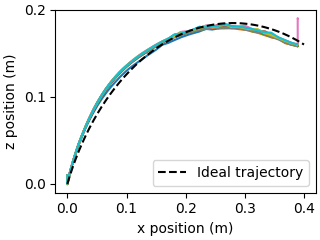}
        \caption{Trajectory, with noise}
        \label{fig:pos_plot_with}
    \end{subfigure}
    \caption{The vertical force and end effector trajectory plots of 20 pivot trials (10 with noise, 10 without). Each trial is shown in a different colour, in addition to the ideal force profile from Equation \ref{eq:pivot_force} and ideal trajectory from Equations \ref{eq:path_x} and \ref{eq:path_z}. The position values are relative to point P as illustrated in Fig. \ref{fig:arc}. The setup uses the long box performing long-to-short pivots using combined modalities.}
    \label{fig:plots}
\end{figure}

To examine the performance and robustness of our position controller in further detail, we plot the wrist force in the z-direction, as well as the pivot path in Fig. \ref{fig:plots}. In Fig. \ref{fig:force_plot_without}, the measured force profile followed the targeted ideal profile closely. The large negative peaks at the beginning and end of the pivot are due to the initial grasp of the box, and slight overshooting which pushed the box into the surface, respectively. It can be seen in Fig. \ref{fig:pos_plot_without} that the pivot could easily adhere to the ideal trajectory in all but  one trial. 

However, when noise is added to the initial planned path, the system can no longer track the force profile as cleanly. In Fig. \ref{fig:force_plot_with}, the measured force profile diverted from the calculated ideal, and large fluctuations are also present. This is due to the frequent vertical adjustments that the gripper needs to make in its path to compensate for the applied noise. As compared to the noise-free case when the gripper moved in a relatively smooth motion, the assumption of constant speed is less valid. Between the peaks, during the stop in motion of the robot arm, the force profiles remain close to ideal predictions. Despite diversions from the ideal force profile, the pivot path as shown in Fig. \ref{fig:pos_plot_with} remained very close to the ideal plan. Influenced by the increased base dimension of the box, an increase to the radius of the path is observed as pivoting begins, corresponding to the large difference in the force plot. The controller then begins acting and the movement converges back to the ideal path, allowing all trials to pivot successfully. If larger forces are involved when handling heavier or larger objects, however, it may be necessary to also consider the dynamic effects of the robot movement, or otherwise implement smoother movement for the gripper.

\subsection{Effects of different objects}
\textbf{Mass:} During our experiments, it was observed that the mass of the box could directly affect the success rate. For the force-based position and vision control methods performing short-to-long pivots, pivoting the large box was always unsuccessful. The failed cases were caused by the low weight of the large box, which is the lightest out of all three boxes. This could not generate enough torque around the gripper fingers to induce rotational slip for the purpose of pivoting. Sufficient torque about the gripper fingers is necessary to create rotational slip as the the box is more capable of overcoming the friction force generated by the grasp. 
On the contrary, the long box showed a 0\% success rate in terms of the pick\&place method. During those experiments, the mass was the main reason for failure as the weight of the box caused too much slip for the robot to rotate the box 90$\degree$.

\textbf{Object shape and Mass distribution:} Besides the mass, the shape of the objects and their mass distributions also affect the gravitational torque around the gripper fingers. The long-to-short pivoting case has a higher torque than the short-to-long case, since the grasp point is further from the centre of mass, roughly at the centre of the box. This helps to induce rotational slip at the grasp point and complete the pivot action. As seen in Table \ref{table:box_1_results}, the success rates for the long-to-short pivots were generally higher for all methods, and the lift rates were generally lower. This was especially significant for the single-modal approaches without gripper control, as they lack the means to regulate the extrinsic influence of gravitational torque by varying gripper width.

However, the iron weights used to manually adjust the weight of the boxes would not have been perfectly distributed around the centre, making the estimation of the torque around the fingers less accurate.

\subsection{Artificial Markers for Pose Estimation}
In the vision component of our system, the ArUco marker plays an important role in detecting the object's pose and further tracking the angle between the surface and the box's lowest edge among all approaches. More generalisable methods for pose estimation may include point cloud clustering and feature matching in the 3D space \cite{guo2021efficient}. Such algorithms can also be used to derive the dimensions of the object. Although this may introduce inaccuracies to the vision component, our tests have shown that by using closed loop approaches, the system can succeed in the face of such inaccuracies.

\subsection{Single vs Multi-modal}
In our combined approach, all three modules are necessary for the "perfect" pivoting. Based on the results in Table \ref{table:all_results}, each single-modal controller was not able to pivot the box with 100\% success rate.
Both vision and force-based position controllers have higher success rate than the gripper module, as our chosen challenge scenario introduced noise to the ideal arc trajectory, which could only be amended by those two controllers. For some specific scenarios (i.e. light weight box), the box can easily be lifted without inducing enough rotational slip, and the gripper controller can be more important to complete pivoting. In the case of vision control, the robot only begins to offset the trajectory after the box has been lifted, causing a high lift rate of 99.2\%.

\section{Conclusions and Future Work} 

We present a closed loop, multi-modal solution utilising vision, force/torque and tactile sensors for manipulating objects via pivoting. The system is able to control the robot to maintain contact with the surface, and modulate the gripper width to induce the desired type of slip. 
% The box is first detected via the ArUco marker attached to one of its faces, which we can then estimate the box's pose. A grasp pose is then generated based off a user input which provides the robot with instructions on which direction to pivot the box. As the gripper closes on the box, it will adjust itself to apply an adequate amount of force to pivot the box. The force controller will then adjust the robot arm by finding the difference of the measured and predicted force at each point throughout its trajectory to determine what adjustment needs to be made. The gripper controller will actively adjust by evaluating if the box is experiencing translational slip or if the gripper is applying excessive force to the box. 

Based on the experiments conducted, we observe a clear advantage of our approach in terms of success rate, robustness, as well as the work done in pivoting the box compared to the open loop and pick-and-place method. The limitations of our approach are its slow execution speed. For future works, we will consider implementing joint velocity control, as well as removing the use of markers for object detection and generalising this approach for many different shapes of objects. Additionally, using learning to perform force-based position and gripper width control will be considered. 

This paper should be considered as a proof-of-concept that demonstrates the feasibility of our pivoting approach by taking advantage of multi-modal sensing. While our work featured only cubic objects tracked via fiducial markers, we believe that our approach can be generalised to rigid objects with modifications. The marker-based pose detection can be swapped with a model-based 6D pose detector \cite{xiang2018posecnn}. A model-based \cite{tremblay2018corl:dope} or model-free grasp synthesis approach \cite{mousavian20196} can be used for obtaining a robust 6D grasp pose. In our work, pivoting on an edge of the box was obvious, however, to generalise it to any given object, further geometric reasoning would be needed for finding a suitable pivot point or edge on the object. Furthermore, a more sophisticated control approach might be needed for contact configuration regulation \cite{doshi2022manipulation}.

\bibliographystyle{ieeetr}
\bibliography{references.bib}
\balance
\end{document}